\documentclass[letterpaper, 10 pt, conference]{ieeeconf}  % Comment this line out if you need a4paper
\usepackage{svg}
\usepackage{float}
\usepackage{stfloats}
\usepackage{placeins}
\usepackage{diagbox}
\usepackage{epsfig} % for postscript graphics files
\usepackage{amsmath} % assumes amsmath package installed
\usepackage{amssymb}  % assumes amsmath package installed
\usepackage{color}
\usepackage[linesnumbered,ruled]{algorithm2e}
\usepackage[normalem]{ulem} %to strike the words
\makeatletter
\let\NAT@parse\undefined
\makeatother
\usepackage{hyperref}
\hypersetup{pdfstartview=FitH,
            colorlinks=true,
            linkcolor=red,
            anchorcolor=blue,
            citecolor=green
            }
\usepackage[table]{xcolor}
\usepackage{cite}
\usepackage{booktabs}
\usepackage{multirow}
\usepackage[caption=false,font=footnotesize]{subfig}

\newcounter{RNum}

\usepackage{pifont}
\usepackage{booktabs}

\usepackage{xcolor}
\definecolor{table_c}{RGB}{241,241,241}
\definecolor{figure_c}{RGB}{241,241,241}
\IEEEoverridecommandlockouts      
                           
\title{\LARGE \bf
Dual Prompt-Driven Feature Encoding for Nighttime UAV Tracking
}
\author{Yiheng Wang$^{1}$, Changhong Fu$^{2,3,*}$, Liangliang Yao$^{2}$, Haobo Zuo$^{4}$, and Zijie Zhang$^{2}$% <-this % stops a space
\thanks{$^{1}$ Yiheng Wang is with the Pratt School of Engineering, Duke University, Durham 27705, United States.}%
\thanks{$^{2}$ Changhong Fu, Liangliang Yao, and Zijie Zhang are with the School of Mechanical Engineering, Tongji University, Shanghai 201804, China.}%
\thanks{$^{3}$ Changhong Fu is with the Shanghai Key Laboratory of Wearable Robotics and Human Machine Interaction, Tongji University, Shanghai 201804, China (e-mail: changhongfu@tongji.edu.cn).}
\thanks{$^{4}$ Haobo Zuo is with the School of Computing and Data Science, The University of Hong Kong, Hong Kong, 999077, China.}
\thanks{* Corresponding author.}
}
\begin{document}
\bstctlcite{IEEEexample:BSTcontrol}
\maketitle
\thispagestyle{empty}
\pagestyle{empty}
%%%%%%%%%%%%%%%%%%%%%%%%%%%%%%%%%%%%%%%%%%%%%%%%%%%%%%%%%%%%%%%%%%%%%%%%%%%%%%%%
\begin{abstract}
% Nighttime UAV tracking has recently obtained substantial performance improvements enabled by prompt tuning techniques.
% Previous prompt tuning approach for nighttime UAV tracking generates darkness clue prompts and refines the representations through gated aggregation of prompt tokens.
% However, the gated aggregation offers only shallow and coarse-grained prompt–representation interaction, preventing representations from fully leveraging the semantic cues provided by the prompts.
% Moreover, the prompter overlooks the view-specificity of UAV tracking, resulting in substantial information loss.
% Robust visual tracking of unmanned aerial vehicles (UAVs) in nighttime environments is of great importance for both research and practical applications.
% However, state-of-the-art (SOTA) methods exhibit poor performance and limited generalization capability in this scenario due to the inherent domain shift from general tracking to nighttime UAV tracking conditions.
% The illumination and viewpoint cues embedded within feature representations play a pivotal role in robust nighttime UAV tracking. 
% However, existing feature encoding approaches largely overlook these critical factors, resulting in inadequate representations that hinder effective adaptation to nighttime UAV tracking.
Robust feature encoding constitutes the foundation of UAV tracking by enabling the nuanced perception of target appearance and motion, thereby playing a pivotal role in ensuring reliable tracking.
However, existing feature encoding methods often overlook critical illumination and viewpoint cues, which are essential for robust perception under challenging nighttime conditions, leading to degraded tracking performance.
To overcome the above limitation, this work proposes a dual prompt-driven feature encoding method that integrates prompt-conditioned feature adaptation and context-aware prompt evolution to promote domain-invariant feature encoding.
Specifically, the pyramid illumination prompter is proposed to extract multi-scale frequency-aware illumination prompts. 
%The dynamic viewpoint prompter adapts the sampling to different viewpoints, enabling the tracker to learn view-invariant features.
The dynamic viewpoint prompter modulates deformable convolution offsets to accommodate viewpoint variations, enabling the tracker to learn view-invariant features.
Extensive experiments validate the effectiveness of the proposed dual prompt-driven tracker (DPTracker) in tackling nighttime UAV tracking. 
Ablation studies highlight the contribution of each component in DPTracker.
Real-world tests under diverse nighttime UAV tracking scenarios further demonstrate the robustness and practical utility.
The code and demo videos are available at \url{https://github.com/yiheng-wang-duke/DPTracker}. 
%\url{https: //github.com/vision4robotics/DPTracker}.
\end{abstract}

%%%%%%%%%%%%%%%%%%%%%%%%%%%%%%%%%%%%%%%%%%%%%%%%%%%%%%%%%%%%%%%%%%%%%%%%%%%%%%%%
\section{Introduction}
%%%% 第一段 %%%%
% UAV tracking has aroused much research attention due to its wide application, \textit{e.g.}, localization and landing~\cite{intro_landing}, traffic monitoring~\cite{intro_monitoring}, and navigation~\cite{intro_navigation}.
% State-of-the-art (SOTA) trackers are oriented for object tracking in general scenarios with fine illumination conditions, and fail to maintain strong performance in adverse nighttime UAV tracking scenarios. The performance drop results from two reasons, \textit{i.e.}, the camera view difference between UAV and general platforms, and the discrepancy of illumination conditions between images captured at nighttime and at daytime.\\
% \indent Previous SOTA prompt tuning-based nighttime UAV tracking method~\cite{dcpt} attempts to tune the foundation model to learn illumination-adaptive representations. It designs the darkness clue prompter alongside the Vision Transformer (ViT). 
% The darkness clue prompters extract the illumination information from the patch embeddings or ViT representations.
% Then the representations are modulated with the summation of selected darkness clue prompt tokens by gated aggregation. 
% However, this method suffers from drawbacks in both representation modulation and prompt generation stages.\\
Unmanned aerial vehicle (UAV) tracking has seen widespread adoption in applications, \textit{e.g.}, localization and landing~\cite{intro_landing}, traffic surveillance~\cite{intro_monitoring}, and autonomous navigation~\cite{intro_navigation}. 
With the advancement of deep neural networks and the availability of large-scale annotated datasets, UAV tracking has achieved remarkable performance gains. 
A key component contributing to these advances is strong
feature encoding ability of deeper models, which enables the construction of discriminative representations by capturing both appearance and motion cues, thereby forming the basis for accurate target localization and trajectory estimation.
However, when applied to nighttime scenarios, UAV tracking remains highly challenging due to severe illumination degradation and low contrast in low-light imagery, as well as the UAV platform’s unique aerial viewpoints. 
The limited consideration of illumination and viewpoint cues in feature encoding of state-of-the-art (SOTA) trackers undermines reliable feature extraction, leading to poor performance in nighttime UAV tracking scenarios~\cite{udat, avtrack}.
% These challenges severely undermine the reliability of feature extraction and tracking performance.
% Therefore, the state-of-the-art (SOTA) trackers underperform in nighttime UAV tracking scenarios for the limited consideration of illumination and viewpoint cues in feature encoding.
Addressing the feature encoding insufficiency is critical for enabling reliable and extensive UAV applications.\\
\begin{figure}[t]
    \centering
    \includegraphics[width=1\linewidth]{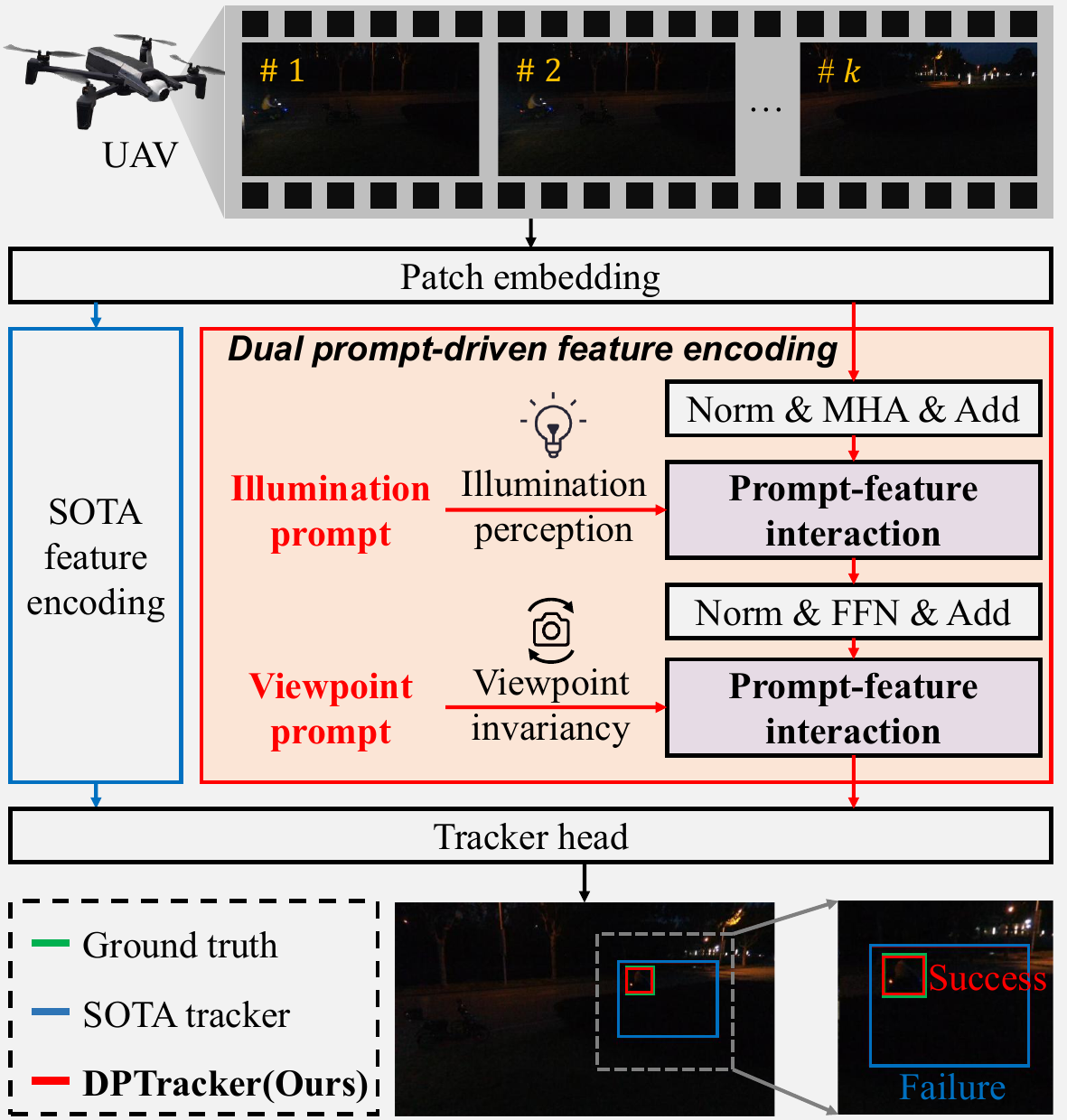}
    \vspace{-20pt}
    \caption{The overall comparison between the SOTA method and the proposed DPTracker. With prompt-feature interaction, DPTracker learns adaptive features and outperforms SOTA trackers in nighttime UAV tracking.}
    \vspace{-30pt}
    \label{fig:fig1 comparison}
\end{figure}
\indent Building on recent advances in deep learning, many SOTA trackers adopt Vision Transformer~\cite{ViT} (ViT) architectures due to the strong capability in modeling global dependencies and spatial relations.
A typical ViT-based tracking paradigm follows a three-stage pipeline: patch embedding, feature encoding, and prediction head. 
Specifically, the input image is first divided into non-overlapping patches, which are then linearly projected into patch embeddings.
The embeddings are processed by a stack of ViT blocks to extract high-level features.
Finally, a tracker head decodes these features to predict the target’s location and bounding box. 
However, the SOTA trackers are not sufficiently effective for nighttime UAV tracking because their feature encoding lacks robustness to domain-specific challenges~\cite{udat}. 
Under low-light conditions and dynamically varying aerial viewpoints, the encoded visual tokens often suffer from degraded quality, misaligned positional cues, and distorted semantic consistency, which collectively lead to biased query–key relationships and unreliable attention formation. 
Therefore, \textit{\textbf{it is essential to incorporate illumination and viewpoint information into the feature encoding stage to ensure reliable representations for robust nighttime UAV tracking}}.\\
% While such architectures have shown promising results under well-illuminated general tracking scenarios, their performance degrades significantly under nighttime UAV tracking conditions, which lack adaptive mechanisms for the weak illumination and distinctive aerial perspectives.\\
% \indent In the representation modulation stage, prompts are introduced to guide the visual representations via a gated aggregation mechanism. 
% However, the modulation process is inherently unidirectional. Prompts influence the visual representations, but the responses of these modulated representations are not utilized to update the prompts in return. As a result, the darkness prompts lack contextual awareness of how the representation space has been modified. This disconnect prevents the prompts from evolving in response to varying representations, leading to suboptimal prompt-representation interaction and limiting the representation adaptability to nighttime UAV tracking scenarios.
% \textbf{\textit{Therefore, designing a bidirectional prompt–representation interaction block is essential to enhance mutual context awareness in prompt update and representation modulation.}}. \\
\indent 
To address the problem, this work proposes the dual prompt-driven feature encoding block (DPBlock) to integrate illumination and viewpoint prompt semantics into the feature encoding of the ViT backbone. 
DPBlock establishes prompt-conditioned guidance, where the generated prompt tokens modulate features, enabling the model to attend to critical visual cues under degraded conditions. 
The prompt-driven feature encoding enhances the model's capacity to adapt representation to low-light environments and dynamic aerial viewpoints, thereby improving robustness in nighttime UAV tracking scenarios.
In parallel, intermediate representations are leveraged to refine the prompt tokens through feedback connections, enabling the prompt tokens to dynamically adjust based on the evolving feature context and better inform subsequent conditioning. \\
\indent This work further designs prompters to learn prompt tokens that capture illumination information and encode the dynamic aerial viewpoint characteristics, as shown in Fig.~\ref{fig:fig1 comparison}. 
The pyramid illumination prompter, inspired by the Laplacian pyramid~\cite{laplacian}, is designed to hierarchically extract multi-scale illumination features from the input image. 
By approximating the Laplacian decomposition structure, the prompter captures frequency-aware features, which are then aggregated into illumination prompt tokens that effectively represent the illumination conditions.
% In addition, a dynamic viewpoint prompter is developed to address the view discrepancy between UAV-mounted and ground-based tracking systems. 
% This module employs deformable convolution to adaptively adjust spatial sampling locations, allowing the viewpoint prompt tokens to capture geometric cues aligned with varying aerial viewpoints and camera poses inherent to UAV platforms.
In addition, a dynamic viewpoint prompter is developed to extract viewpoint information for prompt-driven feature encoding. 
This module leverages deformable convolution to learn adaptive offsets for feature sampling, enabling the viewpoint prompt tokens to perceive geometric variations arising from dynamic UAV viewpoints, thereby enhancing tracker adaptation to diverse aerial viewpoints and camera poses.
The contributions of this paper are summarized as:
\begin{itemize}
\item An innovative dual prompt-driven feature encoding method is proposed,
%wherein illumination and viewpoint prompts act as driving factors that regulate feature encoding and are dynamically optimized through feedback from intermediate representations.
which integrates the prompt-feature interaction mechanism into the ViT backbone, enabling explicit bidirectional adaptation between prompt tokens and visual features for illumination-aware and viewpoint-invariant representation learning.
\item A novel pyramid illumination prompter is designed to approximate the Laplacian pyramid, hierarchically generate multi-scale illumination features, and further aggregate them into the illumination prompt tokens.
\item A dynamic viewpoint prompter is proposed to generate viewpoint prompt tokens, utilizing deformable convolution to adapt spatial sampling to different viewpoints.
\item Extensive experiments on nighttime UAV tracking benchmarks demonstrate the effectiveness of DPTracker, while real-world tests further validate its applicability and superior performance in practical scenarios.
\end{itemize}

\section{Related Work}
\subsection{Nighttime UAV Tracking}
Recent nighttime UAV tracking research is categorized into four paradigms: low-light image enhancement, domain adaptation, training from scratch, and prompt tuning. 
Low-light enhancement approaches~\cite{darktrack2021} employ plug-and-play modules to brighten images before tracking. While boosting visual quality, such preprocessing is loosely coupled with feature encoding and often yields suboptimal representations for tracking. 
Domain adaptation methods aim to bridge the distribution gap between daytime and nighttime images in an unsupervised manner. 
UDAT~\cite{udat} pioneers this approach by employing adversarial training to learn domain-invariant features. 
SAM-DA~\cite{samda} improves sample quality through SAM-based swelling. 
TDA~\cite{tda} incorporates temporal context adaptation into the domain adaptation framework. 
However, without explicitly modeling illumination and viewpoint cues, unsupervised domain adaptation methods often suffer from training instability and limited robustness. 
Training-from-scratch methods directly train models using nighttime object detection annotations. 
DARTer~\cite{darter}  exploits multi-perspective features and dynamic feature activators to achieve better tracking performance, while MambaNUT~\cite{mambanut} employs Vision Mamba architectures with tailored data sampling and loss scheduler. %to learn nighttime-adaptive representations. 
Nevertheless, training-from-scratch approaches are time-consuming and prone to overfitting due to limited nighttime training data, making the learned feature encoding less generalizable. 
Prompt tuning offers a training-efficient and well-performing alternative solution by injecting task-related priors into the feature encoding. 
However, the feature encoding of existing prompt-based trackers insufficiently incorporates illumination and viewpoint information, which in turn restricts their robustness under challenging nighttime UAV tracking conditions.
\subsection{Prompt Tuning for Tracking}
Prompt tuning methods have emerged as an effective paradigm for adapting frozen pre-trained foundation models to various tracking scenarios with minimal parameter overhead. 
ViPT~\cite{vipt} and OneTracker~\cite{onetracker} introduce modality-relevant visual prompts to bridge the domain gap in multi-modal tracking. 
UM-ODTrack~\cite{um-odtrack} further incorporates dense temporal token association and gated modules for adaptive cross-modal representation learning. % across multiple tasks. 
MM-Track~\cite{mm-track} enhances the robustness of multi-modal representations by adaptively transferring discriminative cues while maintaining temporal consistency. % via a memory-inspired adapter. 
MPT~\cite{mpt} proposes a lightweight motion prompt module that encodes long-term trajectory information into the visual embedding space.
%dynamically fusing motion and vision cues through a fusion decoder and adaptive weighting mechanism. 
DCPT~\cite{dcpt} pioneers the prompt tuning techniques in nighttime UAV tracking.
It introduces a darkness clue prompter that injects anti-dark knowledge into a frozen daytime tracker.
%achieving SOTA nighttime UAV tracking performance. 
However, these prompt tuning methods for visual tracking fail to consider both illumination conditions and dynamic aerial viewpoint in feature encoding, resulting in insufficient feature representations for nighttime UAV tracking.
% most methods are predominantly designed for adapting RGB-based tracking methods to general multi-modal scenarios, 
% However, the darkness clue prompter operates at a single scale and single frequency, lacking the sophisticated capability to capture illumination information across multiple scales and frequencies. And it overlooks the unique view inherent to UAV platforms, failing to provide vital guidance to boost the view-invariance of representations. 
Therefore, developing prompters to extract rich illumination and viewpoint semantics is necessary to advance prompt tuning-based nighttime UAV tracking.

\begin{figure*}[t]
    \centering
    \includegraphics[width=1\linewidth]{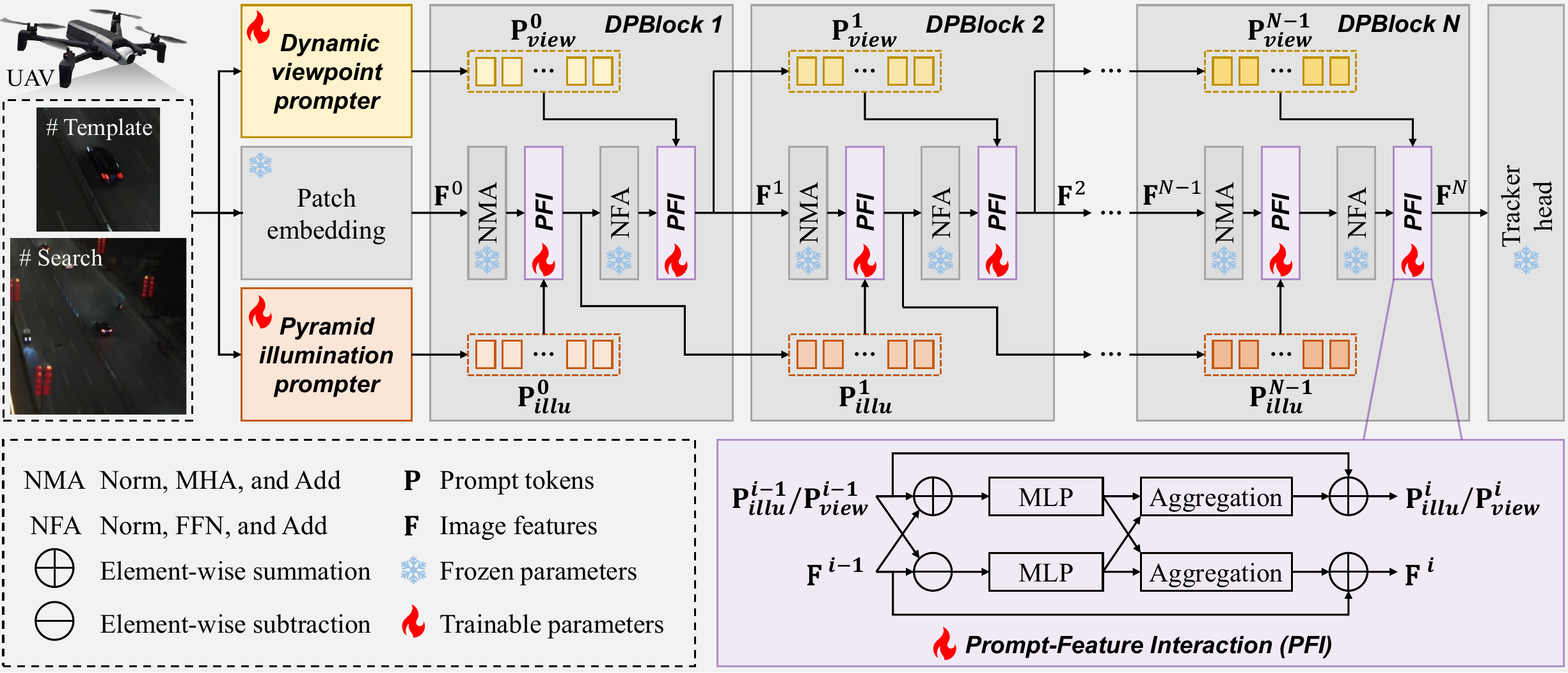}
    \vspace{-20pt}
    \caption{Overall model architecture of DPTracker. 
    The interaction between prompt tokens and features integrates prompt semantics into the features while updating the prompt tokens, thereby enhancing nighttime UAV tracking with more adaptive representations. The images are from NAT2021-\textit{test}~\cite{udat}.}
    \vspace{-20pt}
    \label{fig:fig2 overall}
\end{figure*}
\section{Methodology}
In this section, the proposed method is clearly illustrated.  
Section~\ref{Section:PFI} introduces the dual prompt-driven feature encoding module, which enables prompt-conditioned feature adaptation and context-aware prompt evolution.
Section~\ref{Section:IlluPrompter} presents the pyramid illumination prompter, which extracts multi-scale, frequency-aware illumination prompt tokens.  
Section~\ref{Section:ViewPrompter} describes the dynamic viewpoint prompter, which adaptively deforms convolutional kernels to capture viewpoint prompt tokens from UAV-view images.
%%%%%%%%%%%%%%%%%%%%%%%%%%%%%%%%%%%%%%%%%
%%%%%%%%%%%%%%% 提示-特征交互
%%%%%%%%%%%%%%%%%%%%%%%%%%%%%%%%%%%%%%%%%
\subsection{Dual Prompt-Driven Feature Encoding}
\label{Section:PFI}
To empower feature encoding with illumination and viewpoint cues, this work proposes the dual prompt-driven feature encoding module (DPBlock).
As shown in Fig.~\ref{fig:fig2 overall}, DPBlock integrates two prompt-feature interaction (PFI) modules within the ViT structure.
% The first PFI module is inserted after the self-attention sub-layer of ViT, where illumination prompt tokens are introduced to enhance feature sensitivity to illumination variations, thereby improving the model’s perception of target appearance.
% The second PFI module is placed after the feed-forward sub-layer of ViT to learn viewpoint-invariant representations against dynamic aerial perspective.
The first PFI module is integrated after the self-attention sub-layer of ViT to disambiguate target features from low-contrast backgrounds. By injecting illumination-aware semantics, it compensates for MHA’s tendency to produce indistinguishable responses in poorly-lit environments.
The second PFI module follows the feed-forward sub-layer to cultivate viewpoint-invariant representations. While FFN deepens per-token semantics, the PFI module rectifies the severe geometric distortions inherent in UAV tracking with dynamic viewpoints.
Specifically, PFI is composed of two complementary methods, \textit{i.e.}, prompt-conditioned feature adaptation and context-aware prompt evolution. 
Prompt-conditioned feature adaptation integrates semantic cues from the prompt tokens into the feature representations.
Context-aware prompt evolution updates the prompt tokens with the latest feature context, enabling them to convey more precise guidance for feature adaptation.\\
\indent 
% The prompt-conditioned feature adaptation applies element-wise summation and subtraction to characterize both similarity and difference between the prompt tokens $\mathbf{P}_k^i\in \mathbb{R}^{N\times d}$ and the features $\mathbf{R}^i\in \mathbb{R}^{N\times d}$, where 
% $i$ denotes the index of the DPBlock and $k\in\{illu,view\}$ indicates the type of prompt tokens. 
The prompt-conditioned feature adaptation incorporates element-wise summation and subtraction to characterize the interplay between the prompt tokens $\mathbf{P}_k^i \in \mathbb{R}^{N \times d}$ and the features $\mathbf{F}^i \in \mathbb{R}^{N \times d}$, where 
$i$ denotes the index of the DPBlock and $k\in\{illu,view\}$ indicates the type of prompt tokens. 
Specifically, the summation operation reinforces shared feature manifolds to model the similarity between the prompt and the latent representation, while the subtraction isolates residual, distinctive components to capture the discrepancies between them. 
%The similarity-aware features highlight semantic alignment between prompt tokens and features, while the difference-aware features emphasize the complementary information.
Then, the multi-layer perceptron modules are applied to enhance and exaggerate the encoded semantic similarity and difference.
% The similarity emphasizer focuses on consolidating agreement between prompt tokens and features, thereby reinforcing consistency. 
% In contrast, the difference emphasizer preserves diversity by highlighting non-overlapping features, which can be essential for capturing nuanced contextual or intent-specific signals.
% To refine and enhance these intermediate signals,  the similarity emphasizer and difference emphasizer are applied. 
The process is formulated as:
\begin{align}
\mathbf{E}_{sim}^i &= LN(MLP(\mathbf{F}^{i-1}+\mathbf{P}^{i-1}_k))~,\\
\mathbf{E}^i_{dif} &= LN(MLP(\mathbf{F}^{i-1}-\mathbf{P}^{i-1}_k))~,
\end{align}
where \(\mathbf{E}_{sim}^i\) and \(\mathbf{E}_{dif}^{i}\) denote the similarity-aware features and the difference-aware features in the $i$-index DPBlock, \(MLP\) function as the multi-layer perceptron module, and \(LN\) denotes layer normalization.\\
\indent The prompt-conditioned feature
adaptation further incorporates the similarity cues and suppresses difference cues, ensuring that the feature representations emphasize semantic alignment with the prompt tokens. The adaptation utilizes a simple weighted aggregation strategy, formulated as:
\begin{align}
\mathbf{F}^i &= \mathbf{F}^{i-1} + \alpha_{f} \cdot \mathbf{E}_{sim}^i - \beta_{f} \cdot \mathbf{E}_{dif}^i~,
\end{align}
where \(\alpha_{f}\) and \(\beta_{f}\) are learnable coefficients that determine the contribution of each component in the feature adaptation.\\
\indent The context-aware prompt evolution leverages a similar but asymmetric weighted aggregation method. It attenuates similarity cues and enhances difference cues, enabling the prompt tokens to dynamically adapt to the latest feature context and provide more effective guidance for subsequent feature adaptation. The process is formulated as:
\begin{align}
\mathbf{P}^i &= \mathbf{P}^{i-1} - \alpha_{p} \cdot \mathbf{E}_{sim}^{i} + \beta_{p} \cdot \mathbf{E}_{dif}^i~,
\end{align}
where \(\alpha_{p}\) and \(\beta_{p}\) are learnable coefficients.\\
%for context-aware prompt evolution.\\
\textbf{\textit{Remark 1:}} Dual prompt-driven feature encoding method achieves a dynamic interplay where features absorb semantic guidance from prompt tokens, while prompt tokens evolve based on the latest feature context to provide more precise adaptation guidance.
%Moreover, its computation is efficient by leveraging intermediate computation results, cutting down the computation burdens for UAV platforms.

\begin{figure*}[t]
    \centering
    \includegraphics[width=0.95\linewidth]{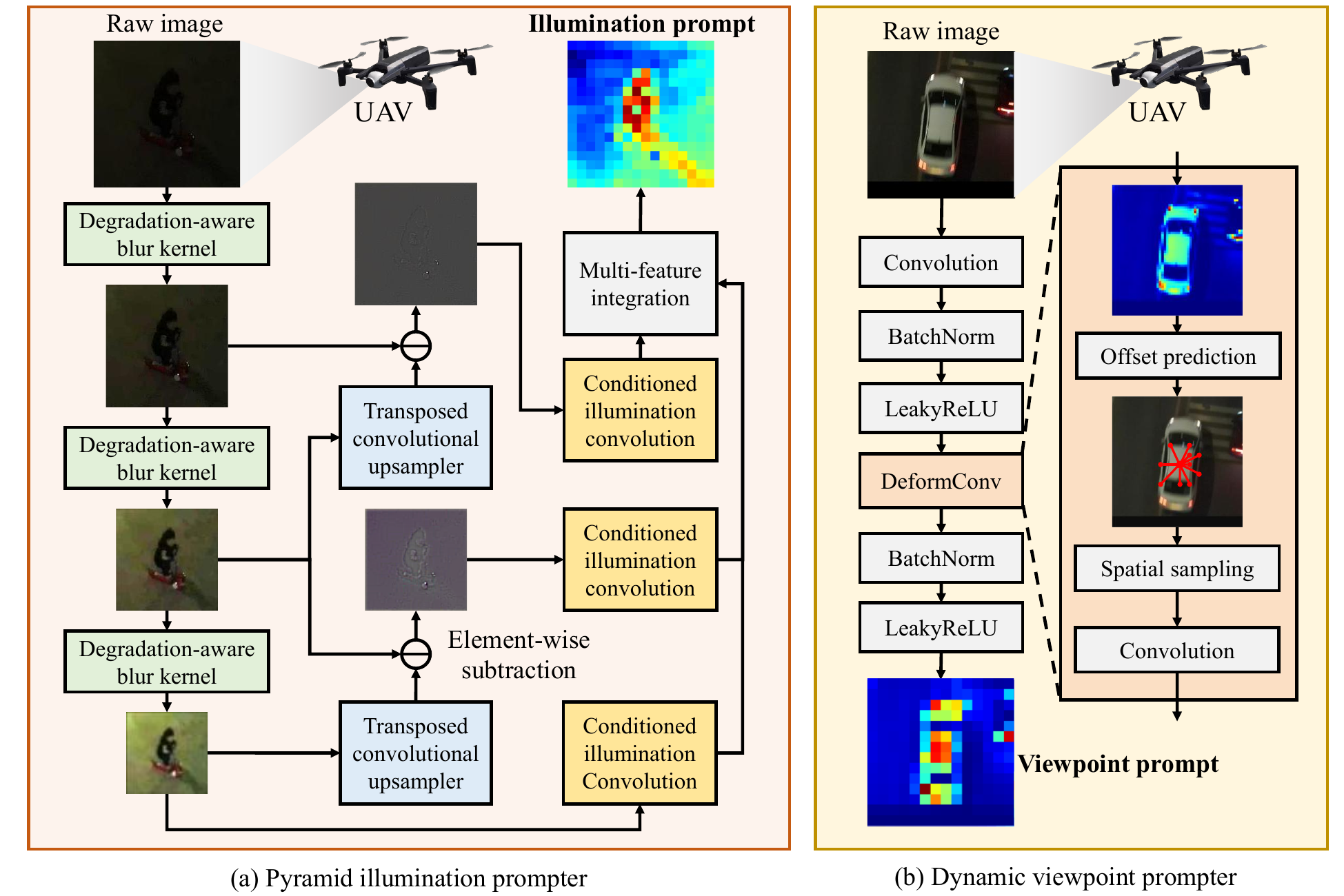}
    \vspace{-10pt}
    \caption{The architecture of the pyramid illumination prompter and the dynamic viewpoint prompter. The illumination prompter leverages a learnable pyramid network to decompose images and aggregate multi-scale features, while the dynamic viewpoint prompter combines standard and deformable convolutions to adaptively capture viewpoint information under UAV perspectives. The images are from DarkTrack2021~\cite{darktrack2021}.}
    \vspace{-20pt}
    \label{fig:fig3 illu prompter}
\end{figure*}

%%%%%%%%%%%%%%%%%%%%%%%%%%%%%%%%%%%%%%%%%%%%%%%%%%%%%%%%%%%%%
%%%%%%%%%%%%%%%%%%%%%% 照度提示器
%%%%%%%%%%%%%%%%%%%%%%%%%%%%%%%%%%%%%%%%%%%%%%%%%%%%%%%%%%%%%
\subsection{Pyramid Illumination Prompter}
\label{Section:IlluPrompter}
To learn nighttime adaptive features, the pyramid illumination prompter (PIP) is proposed to comprehensively encode the lighting conditions into the illumination prompt tokens. 
Previous research~\cite {laplacian_enhancer} has demonstrated that the low-frequency levels of the Laplacian pyramid encode substantial illumination and contrast-related information, while the mid-frequency and high-frequency levels of the Laplacian pyramid focus on structural information. 
Therefore, the Laplacian pyramid is particularly well-suited for learning the illumination prompt.
Building upon this insight, the PIP is designed with a hierarchical architecture to approximate the Laplacian pyramid structure, as shown in Fig.~\ref{fig:fig3 illu prompter}. 
The traditional non-learnable operators for constructing the Laplacian pyramid are replaced by learnable convolutional components to improve the prompt learning capabilities. \\ 
\indent Given a raw image \(\mathbf{I}\), the PIP employs degradation-aware blur kernels whose weights are initialized to approximate Gaussian blur kernels, which preserves the hand-crafted prior knowledge and allows the model to adaptively refine features. 
%These layers are configured with a stride of 2 to perform downsampling.
The PIP constructs a multi-scale feature hierarchy inspired by the Gaussian pyramid, formulated as:
\begin{equation}
    \mathbf{G}_0 = \mathbf{I}~, \quad
    \mathbf{G}_{i+1} = DBK(\mathbf{G}_i)~,
\end{equation}
where \(DBK(\cdot)\) denotes the learnable degradation-aware blur kernel at level \(i\), and \(\mathbf{G}_i\) denotes the blurred image. \\
\indent Traditional upsampling operations are replaced with transposed convolutional upsamplers, which allow the model to learn richer and more informative representations of illumination at each scale. Residual connections are introduced by subtracting the upsampled lower-frequency features from the corresponding higher-frequency features, approximating the Laplacian response at each scale. The whole process is formulated as:
\begin{equation}
    \mathbf{L}_i = \mathbf{G}_i - UP_i(\mathbf{G}_{i+1})~,
\end{equation}
where \(UP_i(\cdot)\) denotes the transposed convolutional upsampler at level \(i\) and \(\mathbf{L}_i\) denotes the corresponding Laplacian component at level \(i\). \\
\indent The conditioned illumination convolutional layers are applied to each Laplacian component, and the final illumination prompt \(\mathbf{P}_{illu}\) is generated by concatenating the responses from all levels along the channel dimension:
\begin{equation}
    \mathbf{P}_{illu}^0 = \mathrm{Concat}(Conv_0(\mathbf{L}_0), 
    % Conv_1(\mathbf{L}_1), 
    \ldots, 
    Conv_n(\mathbf{L}_n))~,
\end{equation}
where \(\mathrm{Concat(\cdot)}\) denotes the channel-wise concatenation and \(n\) is the total number of scales.\\
\textbf{\textit{Remark 2:}} By emulating the Laplacian pyramid in a learnable and differentiable manner, the PIP effectively captures multi-scale illumination information. The illumination prompt tokens are essential for adapting representations to various illumination conditions in nighttime UAV tracking scenarios.

%%%%%%%%%%%%%%%%%%%%%%%%%%%%%%%%%%%%%%%%%%%%%%%%%%%%%%%%%%%%%
%%%%%%%%%%%%%%%%%%%%%% 视角提示器
%%%%%%%%%%%%%%%%%%%%%%%%%%%%%%%%%%%%%%%%%%%%%%%%%%%%%%%%%%%%%
% \begin{figure}[t]
%     \centering
%     \includegraphics[width=1\linewidth]{assets/ViewPrompter-Fig3.pdf}
%     \caption{The detailed architecture of the dynamic viewpoint prompter. It first applies a standard convolutional layer to extract local features from small patches, followed by a deformable convolution layer that performs selective sampling to adaptively capture view information under UAV perspectives.}
%     \label{fig:fig3 viewpoint prompter}
%     \vspace{-10pt}
% \end{figure}
\subsection{Dynamic Viewpoint Prompter}
\label{Section:ViewPrompter}
% Due to the viewpoint discrepancy between ground-level training data and aerial perspective for the nighttime UAV tracking task,
Nighttime UAV tracking suffers from significant geometric and semantic variations due to the inherently dynamic nature of aerial viewpoints.
% The discrepancies remain a critical problem in challenging nighttime UAV target tracking scenarios. 
To address this issue, a viewpoint prompter capable of dynamically capturing geometric information is introduced, thereby facilitating feature adaptation to the dynamic viewpoints of nighttime UAV tracking.\\
\indent As shown in Fig.~\ref{fig:fig3 illu prompter}, the dynamic viewpoint prompter (DVP) is designed to learn viewpoint prompt tokens through a coarse-to-fine prompt generation strategy. 
Initially, DVP employs a standard convolutional layer to extract local features from small image patches, capturing spatial structures that serve as coarse viewpoint prompt tokens of the scene. 
Given an input image \(\textbf{I}\), the coarse viewpoint prompt tokens are computed as:
\begin{equation}
    \textbf{P}_{view}^{c} = LR(BN(Conv(\textbf{I})))~,
\end{equation}
where \(\textbf{P}_{view}^{c}\) denotes the coarse viewpoint prompt tokens, \(BN\) represents the batch normalization, and \(LR\) denotes the LeakyReLU activation function.\\
\indent DVP subsequently applies a deformable convolutional layer to further enhance finer viewpoint adaptivity.
Unlike standard convolutions with fixed grid-based sampling, deformable convolutions introduce learnable spatial offsets, enabling each kernel position to adaptively sample from geometrically relevant regions.
The offset field \(\Delta P \in \mathbb{R}^{2K \times H \times W}\), where \(K\) is the number of sampling points, is predicted from the coarse viewpoint prompt tokens \(\textbf{P}_{view}^{c}\)  by an offset generation module, which is formulated as:
\begin{align}
    \Delta P &= OffConv(\textbf{P}_{view}^{c})~, \\
    \Delta P &= \{\Delta p_1, \Delta p_2, \dots, \Delta p_K\}~, 
\end{align}
where \(\Delta P\) is the predicted offset and \(OffConv\) is the offset convolutional layer. Each \(\Delta p_i \in \mathbb{R}^2\) denotes the offset for the \(i\)-th sampling position.\\
\indent The deformable convolution at position \(p_0\) is computed as:
\begin{equation}
 \textbf{R}(p_0) = \sum_{k=1}^{K} w_k \cdot \textbf{P}_{view}^{c}\big(p_0 + p_k + \Delta p_k\big)~,
\end{equation}
where $p_0$ is the reference spatial location on the feature map,
$p_k$ denotes the predefined sampling location in the regular convolution kernel,
$w_k$ denotes the learnable kernel weight for the $k$-th sampling location,
and $\textbf{R}(p_0)$ is the output feature at location $p_0$.\\
\indent
DVP further applies batch normalization and LeakyReLU activation to generate the final fine-grained viewpoint prompt tokens, which is formulated as:
\begin{equation}
    \textbf{P}_{view}^{0} = LR(BN(\textbf{R}))~,
\end{equation}
where \(\textbf{P}_{view}^0\) denotes the viewpoint prompt tokens.\\
\textbf{\textit{Remark 3:}} 
DVP produces viewpoint prompt tokens that encapsulate rich geometric and semantic cues through coarse-to-fine refinement. The viewpoint prompt tokens provide vital adaptive guidance for feature learning under challenging nighttime UAV viewpoints.

\begin{figure*}[t]
    \centering
\includegraphics[width=1\linewidth]{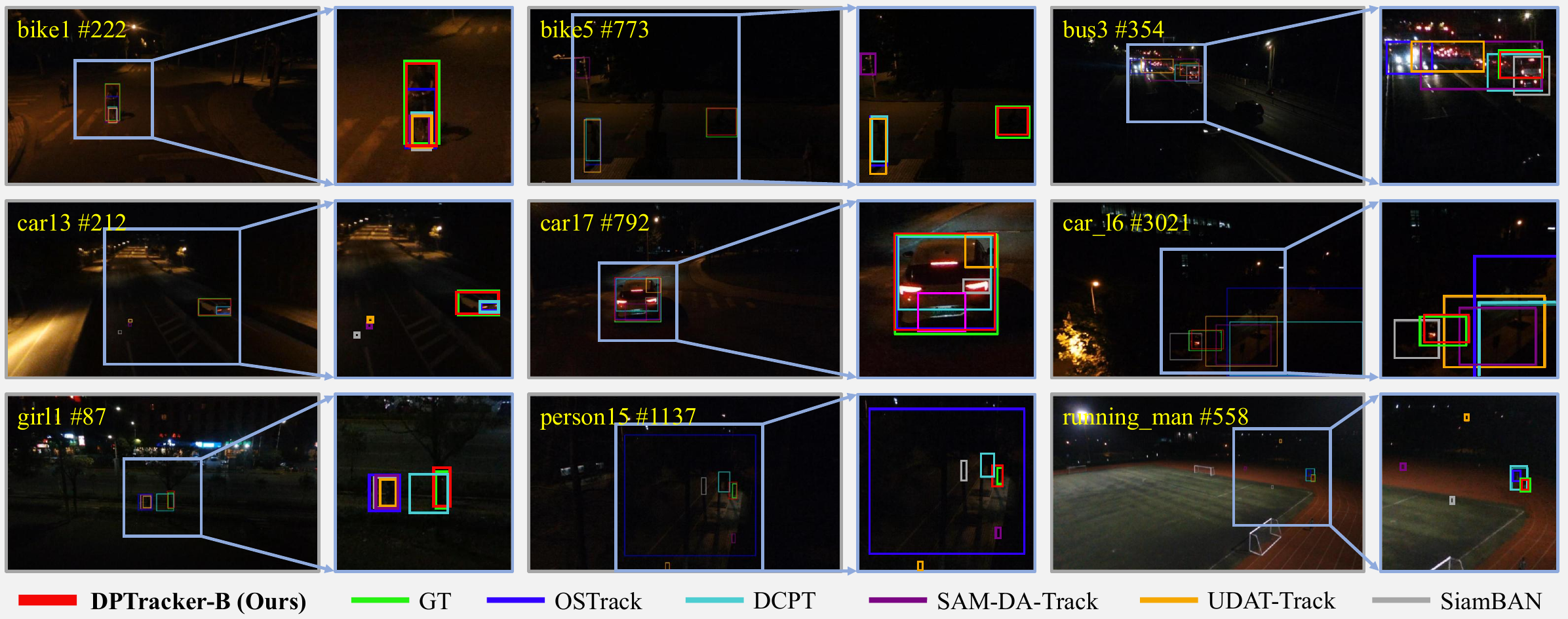}
    \caption{Tracking result visualization of DPTracker-B along with other top trackers~\cite{ostrack, dcpt, samda, udat, siamban}. The sequences are selected from UAVDark135~\cite{uavdark135}. The proposed DPTracker-B shows more robust and precise tracking performance under diverse nighttime UAV tracking scenarios.}
    \vspace{-15pt}
    \label{fig:fig5 visualization}
\end{figure*}
\begin{table*}[b]
\caption{Comparison of tracking performance on UAVDark135~\cite{uavdark135}, DarkTrack2021~\cite{darktrack2021}, and NAT2021-\textit{test}~\cite{udat} benchmarks. The proposed DPTracker-B and DPTracker-T outperform other trackers by notable margins.}
\vspace{-10pt}
\centering
\setlength{\tabcolsep}{4.5pt}
\colorbox{table_c}{
\begin{tabular}{l|l|l|ccc|ccc|ccc|c}
\toprule
\textbf{Method} & \textbf{Source} &
\textbf{UAV} &
\multicolumn{3}{c|}{\textbf{UAVDark135~\cite{uavdark135}}} &
\multicolumn{3}{c|}{\textbf{DarkTrack2021~\cite{darktrack2021}}} &
\multicolumn{3}{c|}{\textbf{NAT2021-\textit{test}~\cite{udat}}} & \textbf{Params} \\
& & &
Prec. & Norm.Prec. & Succ. &
Prec. & Norm.Prec. & Succ. &
Prec. & Norm.Prec. & Succ. & (M) \\
\midrule
% \rowcolor{gray!20}
% \multicolumn{12}{c}{\textbf{Lightweight Tracking Method}} \\
% \midrule
SiamAPN~\cite{siamapn} & ICRA 21 & \ding{51} & 42.4 & 39.5 & 30.1 & 41.9 & 37.9 & 30.7 & 55.8 & 41.8 & 33.7 & 15.2\\
SiamAPN++~\cite{siamapn++} & IROS 21 & \ding{51} & 42.3 & 39.1 & 32.5 & 48.3 & 43.5 & 36.5 & 60.8 & 48.4 & 40.8 & 15.4\\
TCTrack~\cite{tctrack} & CVPR 22 & \ding{51} & 48.8 & 45.4 & 36.2 & 53.5 & 46.8 & 39.3 & 61.2 & 46.8 & 39.4 & 10.4\\
TCTrack++~\cite{tctrack++} & TPAMI 23 & \ding{51} & 47.7 & 44.5 & 37.0 & 56.4 & 48.8 & 42.1 & 61.7 & 48.1 & 40.8 & 15.2\\
AVTrack-ViT~\cite{avtrack} & ICML 24 & \ding{55} & 56.1 & 51.0 & 44.8 & 56.1 & 49.4 & 44.4 & 61.7 & 50.6 & 44.3 & 20.9\\
% % SMAT & WACV 24 & 56.0 & 51.3 & 45.3  & 55.5 & 50.4 & 44.9 & 63.1 & 52.3 & 45.9 & 3.7\\
\textbf{DPTracker-T} & \textbf{Ours} & \ding{51} & \textbf{56.8} & \textbf{51.8} & \textbf{45.6} & \textbf{59.1} & \textbf{51.9} & \textbf{46.5} & \textbf{65.4} & \textbf{53.8} & \textbf{47.5} & 21.6   \\
\midrule
% \rowcolor{gray!20}
% \multicolumn{12}{c}{\textbf{Heavyweight Tracking Method}} \\
% \midrule
SiamBAN~\cite{siamban} & CVPR 20 & \ding{55} & 62.1 & 55.3 & 47.5 & 56.9 & 50.3 & 43.1 & 64.7 & 50.9 & 43.7 & 53.9\\
%SiamPW-RBO & CVPR 22 & 65.0 & 58.3 & 50.7 & 61.4 & 54.4 & 47.7 & 68.7 & 54.5 & 47.6 & 20.0 & 21.6\\
SiamRPN++-RBO~\cite{siamrbo} & CVPR 22 & \ding{55} & 63.3 & 57.6 & 49.5 & 58.5 & 52.9 & 45.2 & 68.2 & 54.8 & 46.9 & 54.0\\
UDAT-BAN~\cite{udat} & CVPR 22 & \ding{51} & 63.3 & 56.6 & 47.9 & 56.4 & 50.0 & 42.1 & 69.4 & 54.6 & 46.9 & 55.1\\
OSTrack~\cite{ostrack} & ECCV 22 & \ding{55} & 67.8 & 61.6 & 55.0 & 66.0 & 58.9 & 52.9 & 72.2 & 60.9 & 53.9 & 92.1 \\
SAM-DA~\cite{samda} & ICARM 23 & \ding{51} & 62.3 & 56.2 & 47.9 & 59.2 & 52.9 & 45.4 & 67.0 & 53.6 & 46.0 & 53.9\\
SmallTrack~\cite{smalltrack} & TGRS 23 & \ding{55} & 65.6 & 52.3 & 44.9 & 56.0 & 49.7 & 42.5 & 65.6 & 52.3 & 44.9  & 53.9\\
DCPT~\cite{dcpt} & ICRA 24 & \ding{51} & 70.3 & 63.9 & 57.1 & 67.1 & 59.9 & 53.7 & 69.4 & 58.4 & 52.0 & 92.9\\
\textbf{DPTracker-B} & \textbf{Ours} & \ding{51} & \textbf{72.3} & \textbf{65.2} & \textbf{58.1} & \textbf{69.8} & \textbf{62.2} & \textbf{55.8} & \textbf{73.9} & \textbf{62.5} & \textbf{55.9} & 94.0 \\
\bottomrule
\end{tabular}
}
\label{tab:tracking_results}
\end{table*}
\begin{figure*}[t]
    \centering
    \colorbox{figure_c}{
        \begin{minipage}[t]{0.975\linewidth}
            \begin{minipage}[t]{0.50\linewidth}
                \centering
                \includegraphics[width=\linewidth]{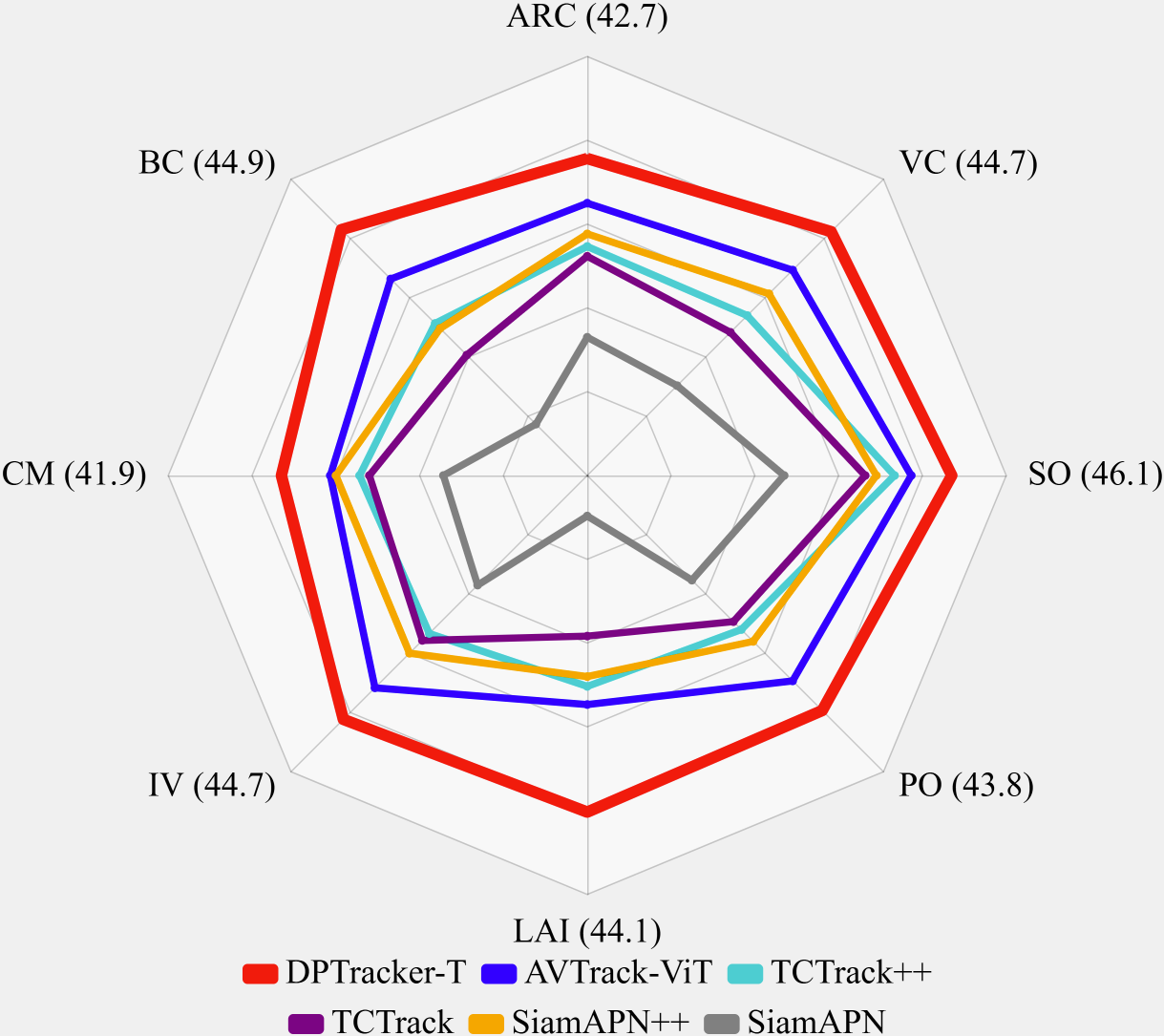}
            \end{minipage}
            \hfill
            \begin{minipage}[t]{0.50\linewidth}
                \centering
                \includegraphics[width=\linewidth]{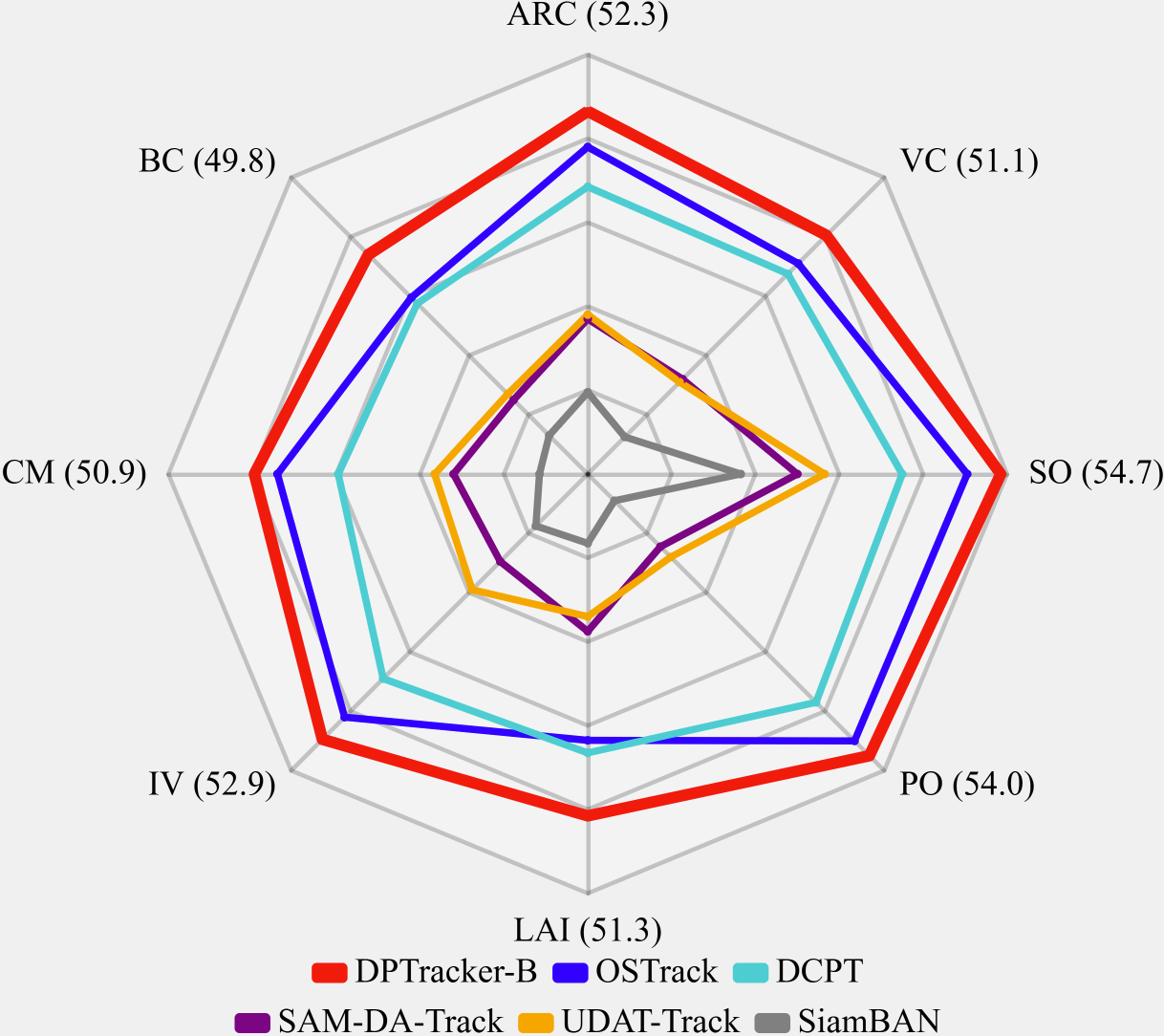}
            \end{minipage}
        \end{minipage}
    }
    \caption{Attribute-based tracking performance evaluation. As shown in the left figure, DPTracker-T consistently achieves the best performance across all 8 attributes on NAT2021-test\cite{udat}, outperforming 5 other lightweight trackers\cite{siamapn, siamapn++, tctrack, tctrack++, avtrack}. In the right figure, DPTracker-B also demonstrates substantial performance improvements compared to other well-performing trackers~\cite{siamban, udat, samda, dcpt, ostrack}.}
    \vspace{-15pt}
    \label{fig:fig4-attribute-evaluation}
\end{figure*}

\section{Experiments}
\subsection{Implementation Details}
The models developed upon existing SOTA general object tracking models AVTrack~\cite{avtrack} and OSTrack~\cite{ostrack} baselines are denoted as DPTracker-T and DPTracker-B, respectively.
A mixture of daytime and nighttime object tracking data are used for training.
The daytime training data include GOT-10K~\cite{got10k}, LASOT~\cite{lasot}, COCO~\cite{coco}, and TrackingNet~\cite{trackingnet}, while the nighttime training datasets include ExDark~\cite{exdark}, Shift~\cite{shift}, and BDD100K~\cite{bdd100k}.
The models are trained with classification and regression losses for 30 epochs using the AdamW~\cite{adamw} optimizer on 2 NVIDIA A100 GPUs. 
The initial learning rate is set to 0.0004, which decays at a rate of 10\% at 24 epochs.

\subsection{Overall Performance Evaluation}
As shown in Table~\ref {tab:tracking_results}, 
% to verify the effectiveness and robustness of the proposed method in diverse nighttime UAV tracking scenarios, 
experiments are conducted to comprehensively evaluate the proposed method against SOTA trackers on three benchmarks~\cite{uavdark135, darktrack2021, udat}. 
The bounding box in the first frame of each sequence is provided.\\
\indent\textbf{UAVDark135}~\cite{uavdark135} is a widely recognized benchmark designed for evaluating object tracking performance in low-light UAV scenarios, comprising 135 video sequences.
Compared with SOTA trackers of similar model size, DPTracker-T achieves the best overall performance across all three metrics, with precision of \textbf{56.8\%}, normalized precision of \textbf{51.8\%}, and success rate of \textbf{45.6\%}, surpassing other SOTA lightweight tracking methods. 
% Compared to previous state-of-the-art methods~\cite{dcpt}, it improves by \texttt{+X.X}, \texttt{+X.X}, and \texttt{+X.X} respectively, demonstrating notable gains in both robustness and accuracy.
As tracker size increases, DPTracker-B establishes new state-of-the-art results with precision of \textbf{72.3\%}, normalized precision of \textbf{65.2\%}, and success of \textbf{58.1\%}.
Compared to the SOTA method~\cite{dcpt}, DPTracker-B improves by \textbf{+2.0\%}, \textbf{+1.3\%}, and \textbf{+1.0\%} in three metrics respectively, demonstrating notable gains in both robustness and accuracy.
As shown in Fig.~\ref{fig:fig5 visualization}, the visualization results indicate the superior capability of DPTracker-B in nighttime UAV tracking.

\textbf{DarkTrack2021}~\cite{darktrack2021} is an authoritative dataset comprising 110 sequences, widely adopted for evaluating UAV tracking under nighttime conditions.
Compared with trackers of similar size, DPTracker-T achieves superior performance across all three evaluation metrics, reaching \textbf{59.1\%} precision, \textbf{51.9\%} normalized precision, and \textbf{46.5\%} success rate, exceeding all competing tracking methods.
When scaling to larger trackers, DPTracker-B outperforms other models with precision \textbf{69.8\%}, normalized precision \textbf{62.2\%}, and success rate \textbf{55.8\%}. Compared to the SOTA tracker~\cite{dcpt}, DPTracker-B achieves a relative improvement of \textbf{+2.7\%}, \textbf{+2.3\%}, and \textbf{+2.1\%} across the three metrics, underscoring its robustness and strong adaptability in nighttime UAV tracking.

\textbf{NAT2021-\textit{test}}~\cite{udat} is a large-scale benchmark containing 180 video sequences.
%, designed to evaluate aerial tracking performance in diverse nighttime conditions.
When evaluated against trackers of similar scale, DPTracker-T achieves new state-of-the-art performance with precision of \textbf{65.4\%}, normalized precision of \textbf{53.8\%}, and success rate of \textbf{47.5\%}.
As the trackers scale up, DPTracker-B performs the best among all trackers with precision \textbf{73.9\%}, normalized precision \textbf{62.5\%}, and success rate \textbf{55.9\%}. These consistent improvements confirm the effectiveness of the proposed method in nighttime UAV tracking across various challenging scenarios.
\subsection{Attribute-Based Evaluation}
% \begin{figure}[t]
%     \centering
%     \includegraphics[width=1\linewidth]{assets/heavy.png}
%     \includegraphics[width=1\linewidth]{assets/light.png}
%     \caption{Attribute-based evaluation of 6 top trackers (Need to update).}
%     \label{fig:fig4 attribute evaluation}
% \end{figure}
To further investigate tracking robustness under diverse nighttime UAV tracking challenges, a further attribute-based evaluation, as shown in Fig.~\ref{fig:fig4-attribute-evaluation}, is conducted across eight key factors: aspect ratio change (ARC), background clutter (BC), camera motion (CM), illumination variation (IV), low ambient intensity (LAI), partial occlusion (PO), similar object (SO), and viewpoint change (VC).
DPTracker-T achieves the most robust performance among trackers with similar model scales, excelling under low ambient intensity \textbf{44.1\%} and illumination variation \textbf{44.7\%}. 
DPTracker-B also demonstrates remarkable superiority under illumination-related challenges, achieving the highest success rates in scenarios with low ambient intensity \textbf{51.3\%} and illumination variation \textbf{52.9\%}. 
%clearly surpassing existing state-of-the-art methods. 
Beyond illumination robustness, DPTracker also consistently outperforms other trackers across the remaining attributes with notable advantages.
These results validate the effectiveness of the proposed approach in handling challenges in nighttime UAV tracking scenarios.

\subsection{Ablation Study}
To investigate the effectiveness of each component, ablation experiments are conducted on PFI, PIP, and DVP using two evaluation metrics: precision and success rate, as shown in Table~\ref{tab:ablation}.
To validate the PFI module in the prompt-driven feature encoding, the prompters are replaced with the patch embedding modules. 
Using PFI brings noticeable relative gains of +5.3\% and +4.5\% in the two metrics, 
demonstrating that PFI can effectively enhance the feature encoding for nighttime UAV tracking. 
PIP and DVP provide granular refinements to further calibrate and polish the learned representations.
Utilizing PIP further achieves a performance of 71.9\% and 57.8\% on the two metrics, highlighting its ability to provide multi-scale frequency-aware illumination prompts.
% which guide the tracker to learn more reliable feature representations under various light conditions.
Moreover, incorporating DVP yields a performance of 72.3\% and 58.1\% on the two metrics, demonstrating enhanced robustness by enabling the model to capture viewpoint-invariant features across aerial perspectives.
These results verify that PFI, PIP, and DVP jointly contribute to robust nighttime UAV tracking.
\begin{table}[!b]
 \vspace{-10pt}
  \centering
  \caption{Ablation study of the PFI, PIP, and DVP. The performance on UAVDark135 verifies the contribution of each component.}
  \vspace{-5pt}
  \colorbox{table_c}{
  \begin{tabular}{ccc|cc|cc}
  \toprule
  PFI & PIP & DVP & Prec.(\%) & $\Delta$ (\%) & Succ.(\%) & $\Delta$ (\%) \\
  \midrule \ding{55} & \ding{55} & \ding{55} & 67.8 & - & 55.0 & - \\
  \midrule
  \ding{51} & \ding{55} & \ding{55} & 71.4 & +5.3 & 57.5 & +4.5 \\
  \ding{51} & \ding{51} & \ding{55} & 71.9 &  +6.0  & 57.8 & +5.1 \\
  \midrule
  \ding{51} & \ding{51} & \ding{51} & \textbf{72.3} & \textbf{+6.6} & \textbf{58.1} & \textbf{+5.6} \\
  \bottomrule
  \end{tabular}}
  \label{tab:ablation}%
\end{table}
\begin{figure}[t]
    \centering
\includegraphics[width=1\linewidth]{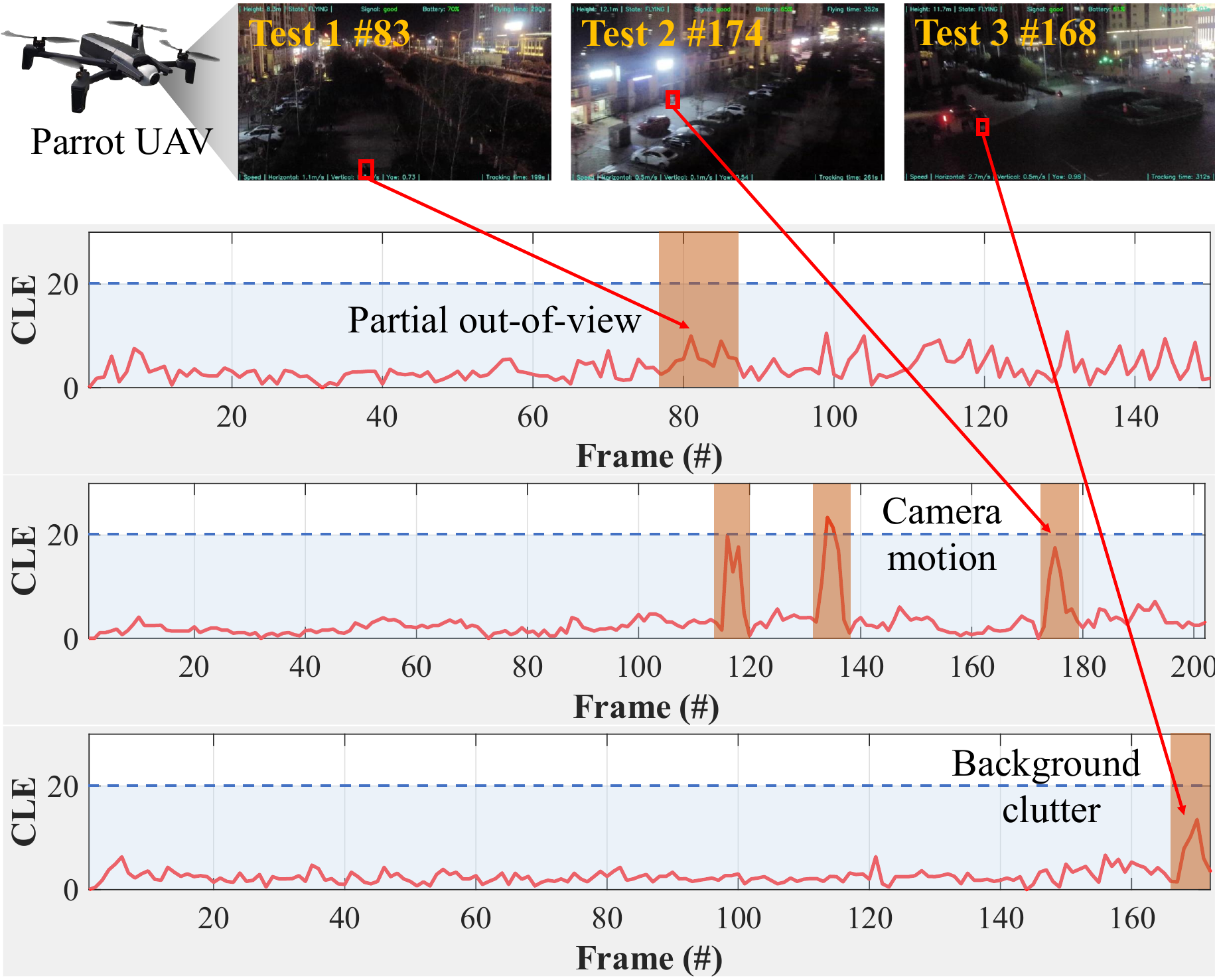}
    \caption{Real-world UAV tracking tests under nighttime conditions.
    Regions in blue, where the CLE is less than 20, are regarded as successful tracking cases, while the orange regions represent challenging tracking conditions.}
    % The proposed DPTracker-T achieves high computational efficiency, operating at fast speeds while consistently maintaining robust target localization under challenging conditions.}
    \vspace{-20pt}
    \label{fig:real test}
\end{figure}
\subsection{Real-World Tests}
To assess real-world performance, a workstation equipped with an NVIDIA RTX 3080Ti GPU serves as the ground control station (GCS). The Parrot UAV captures images at 10 frames per second and transmits them to the GCS via WiFi. 
%During field trials, the bounding box in the first frame is manually given, and then the tracker achieves a processing speed of \textbf{54} frames per second, predicting target positions and transmitting them back to the UAV. 
Upon manual initialization of the target bounding box in the initial frame, the tracker maintains a real-time inference speed of \textbf{54} FPS, facilitating continuous target localization and transmitting target positions back to the UAV.
As presented in Fig.~\ref{fig:real test}, tracking performance is evaluated using the center location error (CLE), where values below 20 are regarded as successful localization. The proposed DPTracker-T demonstrates reliable tracking across diverse nighttime conditions.
Test 1 is conducted under low ambient illumination, where the partial out-of-view of the target further compounds the tracking difficulty.
Test 2 is primarily challenged by camera motion, which introduces significant trajectory variations and appearance blur.
Test 3 involves a street-crossing scene characterized by background clutter. 
The proposed tracker demonstrates robust tracking performance under the challenging nighttime scenarios.

\section{Conclusions}
This work proposes a dual prompt-driven feature encoding method to learn illumination and view-adaptive representations for nighttime UAV tracking. 
Prompt–feature interaction plays a central role in reinforcing the mutual adaptation between prompts and visual features for enhanced representation robustness. 
Built upon this interaction method, the pyramid illumination prompter and dynamic viewpoint prompter further strengthen the model’s ability to accommodate complex illumination conditions and dynamic viewpoint changes.
%By harnessing the multi-scale frequency insights of the Pyramid Illumination Prompter alongside the geometric modeling of the Dynamic Viewpoint Prompter, the model establishes a new paradigm for robust representation learning against extreme illumination and perspective shifts.
Comprehensive experiments and ablation studies verify the effectiveness of these components.
Future research will extend this adaptive prompting paradigm to broader adverse scenarios in autonomous aerial systems.
In summary, this study contributes to low-light object tracking for unmanned aerial vehicles in adverse conditions.
% \section*{Acknowledgement}
% This work was supported in part by the National Natural Science Foundation of China under Grant 62173249 and Grant U24B20161, and in part by the Natural Science Foundation of Shanghai under Grant 20ZR1460100.
\bibliographystyle{IEEEtran}
\bibliography{ref}
\end{document}